 \documentclass[pmlr,twocolumn,10pt]{jmlr} 

\usepackage{caption}
\usepackage{array}
\usepackage{colortbl}
\usepackage{multirow}
\usepackage{amssymb}
\usepackage{placeins}
\usepackage{float}
\usepackage{bm}
\usepackage{amsmath}
\usepackage{enumitem}
\usepackage{float}
\usepackage{afterpage}
\usepackage{graphicx}
\usepackage{xcolor}
\usepackage{booktabs}
\usepackage{siunitx}

\theorembodyfont{\upshape}
\theoremheaderfont{\scshape}
\theorempostheader{:}
\theoremsep{\newline}

\jmlrvolume{LEAVE UNSET}
\jmlryear{2023}
\jmlrsubmitted{LEAVE UNSET}
\jmlrpublished{LEAVE UNSET}
\jmlrworkshop{Machine Learning for Health (ML4H) 2023} 

 \title[Short Title]{ML4H 2023 Template: Proceedings Track}

 \title{\textbf{\fontsize{14}{16} SurvTimeSurvival: Survival Analysis On The Patient With Multiple Visits/Records}}

\author{%
\Name{Hung, Le Nhat} \Email{hl01600@surrey.ac.uk}\\
\addr University of Surrey, United Kingdom
\AND
\Name{Ong, Eng-Jon}\Email{e.ong@surrey.ac.uk}\\
\addr University of Surrey, United Kingdom
\AND
\Name{Bober, Miroslaw} \Email{m.bober@surrey.ac.uk}\\
\addr University of Surrey, United Kingdom
}

\begin{document}

\maketitle


\begin{abstract}
\noindent\emph{
The accurate prediction of survival times for patients with severe diseases remains a critical challenge despite recent advances in artificial intelligence.
This study introduces "SurvTimeSurvival: Survival Analysis On Patients With Multiple Visits/Records," utilizing the Transformer model to not only handle the complexities of time-varying covariates but also covariates data. We also tackle the data sparsity issue common to survival analysis datasets by integrating synthetic data generation into the learning process of our model.
We show that our method outperforms state-of-the-art deep learning approaches on both covariates and time-varying covariates datasets. Our approach aims not only to enhance the understanding of individual patient survival trajectories across various medical conditions, thereby improving prediction accuracy, but also to play a pivotal role in designing clinical trials and creating new treatments. 
\footnote{The source code associated with the methodologies and results presented in this manuscript will be made publicly available and can be accessible at: \href{https://github.com/davidlee1102/Surtimesurvival}{https://github.com/davidlee1102/Surtimesurvival}.} 
}
\end{abstract}

\section{Introduction}
\label{sec:Introduction}

Survival analysis, called time-to-event, is concerned with predicting event occurrence times and is indispensable in medicine \cite{friedman2015fundamentals}, social sciences \cite{box2015survival}, and economic \cite{davidson2019lifelines}, enabling researchers to compute survival probabilities and hazard functions and compare diverse groups or conditions. Traditional models such as the \cite{cox1972regression} and Kaplan-Meier \cite{bland1998survival} estimators have foundational significance but are limited due to the restrictive assumption made. In particular, these limitations are relaxed on recent AI-based solutions like DeepSurv \cite{katzman2018deepsurv}, Deep-Hit \cite{lee2018deephit}, Dynamic Deep-Hit \cite{lee2019dynamic}, and Deep Survival Machines \cite{nagpal2021deep} symbolize evolution in this domain. Notably, the SurvTrace model \cite{wang2022survtrace}, employing deep learning with the Transformer-based architecture \cite{vaswani2017attention, devlin2018bert}, has shown promising predictive accuracy. However, some limitations still remain, affecting survival prediction accuracy.
Specifically, it appears that SurvTrace does not adequately incorporate time-varying covariates (time-cov) in each patient's records, potentially impacting the robustness of this method. On another front, Dynamic Deep-Hit's dependence on LSTM architecture \cite{hochreiter1997long} may constrain its predictive prowess due to a "long-term dependency problem" \cite{zhao2020rnn}.



\textbf{Contribution}. In this work, we propose \textbf{\textit{"SurvTimeSurvival: Survival Analysis On The Patient With Multiple Visits/Records"}} deep-learning with the Transformer-based method that not only leverages the SurvTrace strength but also utilizes time-varying covariates adaptation to more accurately model and comprehend each individual patient's data, enhancing the precision of survival predictions. A more granular exposition of SurvTimeSurvival is elaborated on three different points:

\begin{enumerate}[label=(\alph*)]
\item We develop a new approach that facilitates the management of both covariates and time-varying covariates data by utilizing the advantages of Transformers architecture \cite{vaswani2017attention} with multiple related visits/records, thereby augmenting the comprehension of dependent visits/records that relate to each identified patient.

\item We design a new architecture that combines our proposed model (a) with SurvTrace \cite{wang2022survtrace} method to not only optimize learning across a variety of data and emphasize essential features that influence patient survival duration but also utilize the advantages of SurvTrace.

\item We proposed \textit{SurvTimeSurvival}, a novel approach that strategically combines our integration model (b) with the SurvivalGan technique \cite{norcliffe2023survivalgan} to harness the benefits of synthetic data during training, thereby enhancing survival prediction accuracy.

\end{enumerate}



We have performed comprehensive experiments on \textit{SurvTimeSurvival} using three different public time-varying covariates and covariates medical datasets. The resulting experimental results demonstrate that our method markedly surpasses the state-of-the-art baselines.


\section{Related Work}
\label{sec:relatedwork}

In longitudinal survival analysis in healthcare, "time-varying covariates" (time-cov) encapsulate dynamic changes to patient data across visits/records, improving comprehension of evolving health trends and event risks, such as relapse or death. Contrasted with "time-varying covariates", the covariates data is "non-time-varying covariates" data or "time-invariant covariates" data, meaning each patient has one record/visit only. Time-varying covariates data helps to reveal intricate temporal relationships, enhancing prediction accuracy and clarifying key risk factors \cite{pourjafari2022survival}. 



The Cox PH model \cite{cox1972regression} is fundamental in survival analysis in biomedical research, clarifying how covariates affect survival. However, its assumptions, such as expecting proportionality and linear relationships among covariates, might limit its usefulness in situations where there are complex non-linear interactions. Contemporary methodologies like \cite{katzman2018deepsurv, nagpal2021deep, lee2019dynamic, lee2018deephit} offer advanced analytical prowess, although they are not immune to pitfalls. The prevalent use of RNN \cite{schmidt2019recurrent} and LSTM \cite{hochreiter1997long} networks may not be suited for survival analysis since the "long-term dependency problem" \cite{zhao2020rnn}.


The Transformer architecture \cite{vaswani2017attention} and BERT \cite{devlin2018bert} have improved sequential medical data processing \cite{hu2021transformer}, with BERT using pre-trained models for bidirectional clinical contexts. Moreover, SurvTRACE \cite{wang2022survtrace} is a Transformer-based model integrating numerical embedding, attentive encoding, and an alignment module for tasks like survival analysis, mortality prediction, and length-of-stay prediction. It uses discrete-time hazard rates, PCH loss, and IPS-based PCH loss to improve the prediction of survival outcomes. However, SurvTrace just only deals with covariates (cov) in each patient's records, potentially impacting the results' robustness. Based on that, our research capitalizes on Transformer features for time-varying covariates adaptation handling, optimizing feature interpretation, and enriching SurvTrace strength.

Synthetic data \cite{lu2023machine}, created to mimic real-world data statistically, helps in maintaining patient privacy. SurvivalGAN \cite{norcliffe2023survivalgan} is a GAN \cite{creswell2018generative} variant sculpted capable of synthesizing time-to-event data, crucial for survival analysis— a domain concentrating on durations preceding pivotal events. By being implemented in the Synthcity framework \cite{qian2023synthcity}, SurvivalGAN engenders datasets mirroring authentic data characteristics but sans personal identifiers. Thus, synthetic data is pivotal in tackling the training challenges arising from data insufficiency, a situation often precipitated by privacy issues and technology limitations.

\begin{figure*}[!htb]
\centering
    \includegraphics[width=1\textwidth]{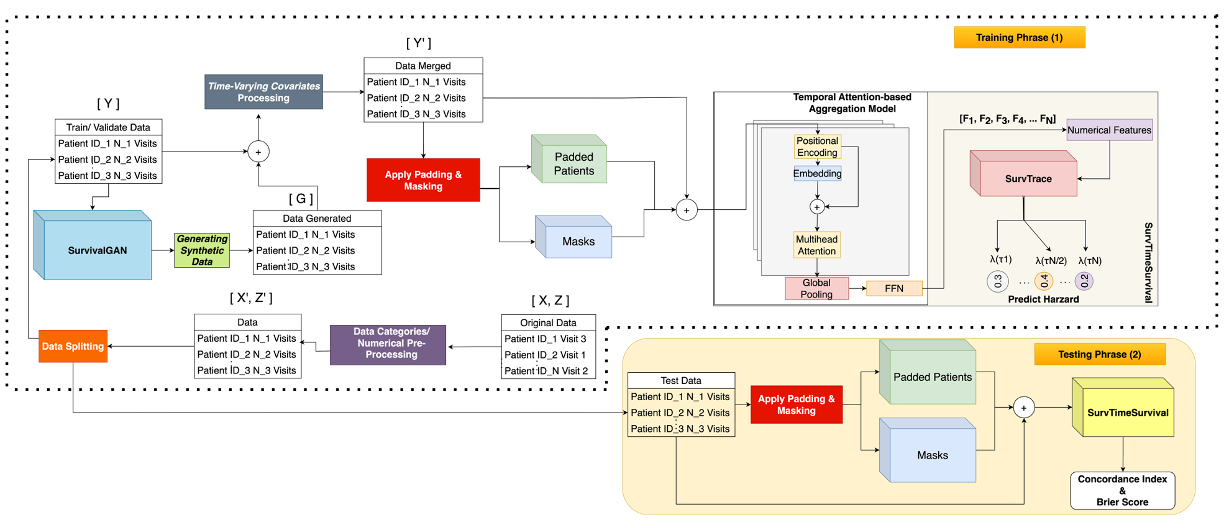}
    \caption{In the SurvTimeSurvival framework, raw data comprising both categorical and numerical covariates is encoded and partitioned into training and test sets. The training set is augmented with synthetic data generated by SurvivalGan \cite{norcliffe2023survivalgan} and subsequently processed to adapt time-varying covariates structure. By adding the masking, it allows the attention mechanism of our model to ignore the padded components of our input. After implementing masking and padding, the data is fed into the SurvTimeSurvival framework, a hybrid model that integrates our Transformer-based model \cite{vaswani2017attention}—designed to handle both data types—and the SurvTrace module \cite{wang2022survtrace}, which utilizes \(N\) features from our module for survival outcomes predictions. The model's performance is then evaluated on the test data. This evaluation is conducted after the data has been processed to adapt time-varying covariates and apply masking and padding functions (phase 2). The evaluation metrics used for this assessment are the Concordance Index and the Brier score.}
\end{figure*}

\section{Model Architecture}
\label{sec:modelarchitecture}

An overview of our SurvTimeSurvival architecture is shown in Figure 1. Its architecture consists of three main components: "Our Transformer-based temporal aggregation method", "SurvTrace
", and "SurvivalGan". 

Specifically, the combination of our temporal aggregator and SurvTrace allows our model to deal with two types of survival analysis data (time-varying covariates and covariates) whilst simultaneously leveraging the strength of SurvTrace \cite{wang2022survtrace}.  Moreover, the integration of SurvivalGan \cite{norcliffe2023survivalgan} for synthetic data generation into our learning process allows us to tackle the issue of data sparsity in survival analysis data. We show in Section \ref{sec:experiments} that this results in a noticeable improvement in the accuracy of survival predictions.

\subsection{Problem Definition}

We define the structure of our survival data as follows:
\[ D = \{(\bm{X}_i,\bm{Z}_i, \bm{t}_i, e_i)\}_{i=1}^{n} \]
The total number of patients is $n$. Each patient (say $i^{th}$ patient), has $N_i$ number of records, where \( \bm{X}_i \in \mathbb{R}^{N_i \times n_{num}} \) represents the \( n_{num} \) {\em numerical } covariate features, \( \bm{Z}_i \in \mathbb{Z}^{N_i \times n_{cat}} \) represents the \( n_{cat} \) {\em categorical } covariate features. The vector of event times is given as \( \bm{t}_i \in \mathbb{R}^{N_i} \).
The indicator \( e_i \) differentiates whether \( \bm{t}_i \) is an actual event or censored. For single events, \( e_i \) is binary, but for competing events, \( e_i \in \{0, 1, \ldots, K_E\} \) specifies the event at \( \bm{t}_i \). A value of \( e_i = 0 \) denotes right-censoring for the \( i \)-th patient, meaning no events were observed by the study's end \cite{wang2022survtrace}.  Note that in the event of covariates data (i.e. non time-varying covariates), we set $N_i = 1$.
%
Central to survival analysis is understanding the hazard and survival functions, with the survival function \( S(t) \) representing the probability a subject is event-free until time \( t \), defined as \( S(t) \equiv \Pr(T > t) \). Meanwhile, the hazard function, symbolized as \( \lambda(t) \), provides a momentary rate at which events transpire, conditional on the absence of prior events. It is defined by the following limit:
\[ \lambda(t) \equiv \lim_{{\Delta t \to 0}} \frac{\Pr(t \leq T < t + \Delta t \,|\, T \geq t)}{\Delta t} \]

The hazard function, \( \lambda(t) \), fundamentally represents the instantaneous risk of event occurrence at time \( t \), given that the subject has remained event-free up to and including time \( t \). In a similar vein, the probability mass function (PMF) associated with the event time is denoted as \( g(t) \), where \( g(t) = \Pr(T = t) \) explicitly characterizes the probability that the event transpires precisely at time \( t \).

\subsection{Survival Data: An Overview of Processing \& Generating Synthetic Data}

Our model follows the common pre-processing step of normalizing numerical features and encoding categorical features. 
With the numerical data \( \bm{X} = \{((x_j)_{j=1}^{n_{num}})_i\}_{i=1}^{n}\) with \( x_j \in \mathbb{R} \), we normalize as \( \bm{X}' = \{((x_j' = \frac{x_j}{M})_{j=1}^{n_{num}})_i\}_{i=1}^{n}\) where \( M \) is \( \bm{X} \)'s maximum. For categorical data \( \bm{Z} = \{((z_j)_{j=1}^{n_{cat}})_i\}_{i=1}^{n}\) with \( z_j \in \mathbb{N} \) undergoes one-hot encoding \cite{xu2019modeling}, producing \( \bm{Z}' = \{((O(z_j))_{j=1}^{n_{cat}})_i\}_{i=1}^{n}\). The unified dataset \( \bm{Y} \) combines \( X' \) and \( Z' \), represented as \( \bm{Y} = \{\left((x_1', \ldots, x'_{n_{num}}, O(z_1), \ldots, O(z_{n_{cat}})\right)_i\}_{i=1}^{n}\), with \( n\) being total visits/records and \(n_{num},  n_{cat}\) are the number of numerical and category features.

After the above pre-preprocessing, the dataset is partitioned into a part for training/validation and the remaining as the test subset. We have ensured that the test subset is not involved in either the model training nor the synthetic data generation. 
Subsequently, the training data undergoes synthesis using the \textit{SurvivalGan} paradigm \cite{norcliffe2023survivalgan}. This framework leverages the power of the Generative Adversarial Network (GAN) \cite{creswell2018generative}, facilitating the proficient generation of synthetic covariates data. The main target of SurvivalGAN is finding the result of the "optimism" equation. Mathematically, its formula is delineated:

\begin{equation*}
\text{Optimism} = \int_0^\infty f(t) [p_{\text{Syn}}(t) - p_{\text{Real}}(t)] \, dt.
\end{equation*}

Where \(f(t)\) is defined as a function for piecewise. If the time interval \(0 \leq t \leq T\), \(f(t)\) takes on the value of \(t/T\), while times greater than \(T\), \(f(t)\) is simply 1. Moreover \(p_{\text{Syn}}(t)\), and \(p_{\text{Real}}(t)\) represent the survival function of the synthetic data and the real data at time \(t\), respectively. Thus, the measure "optimism" quantifies deviations between synthetic and real survival datasets. This promotes a data representation, counteracting biases and enhancing model reliability and prediction accuracy across diverse scenarios. 


Post-generation of synthetic data, we add it to the previous training/validation set. The combined data is then transformed to a time-varying covariates format. Should the data already be in time-varying covariates format, each patient's series of visits/records are amalgamated into a structured sequence, akin to sequence processing approaches in NLP \cite{devlin2018bert, liu2019roberta}. Moreover, let \(\bm{\textit{ST}}\) represent the synthetic feature; the generated and original data will be represented as follows:
\begin{align*}
\bm{G} &= \left\{ \left( \{(O(z_{ST_{v}}))\}_{v=1}^{n_{cat}}, \{(x_{ST_{j}}')\}_{j=1}^{n_{num}}\right)_i \right\}_{i=1}^{m} \\
\bm{Y} &= \left\{ \left( \{(O(z_{v}))\}_{v=1}^{n_{cat}}, \{(x_{j}')\}_{j=1}^{n_{num}} \right)_i \right\}_{i=1}^{n}
\end{align*}

where \( m\) is the number of visits/records generated, with \( n\) being the total visits/records in original data and \(n_{num}, n_{cat}\) is the number of numerical and category features respectively, encompassing both original and synthesized data. We denote the combined dataset of data in $\bm{G}$ and $\bm{Y}$ as $\bm{Y}'$.

Moreover, given a maximum number of visits/records of the patient is \(V \), the number of visits/records equal to V truncates to the \(V\) visits/records, while those with fewer V will be padded as \(<\textbf{{pad}}>\) for uniformity. Specifically, for patient \( \text{id\_1} \) with \( n < V \) visits/records, the sequence is represented as \( \text{id\_1\_sequence} = \{ \text{{\textbf{v}}}_i \}_{i=1}^n \cup \{ <\textbf{{pad}}> \}_{i=n+1}^{V} \), where \( \mathbf{v}_i \) is the \( i \)-th visit/record, containing \(n_{num}, n_{cat}\) features at time \(i\).


\subsection{Temporal Attention-based Aggregation Model}
In this section, we describe our temporal aggregation design that leverages the strength of the Transformer architecture to deal with two different data types: time-varying covariates data and covariates data. Our design is split into 3 main parts: Input representation \& Positional encoding, Transformer encoder layer, and Features extraction through the embedding layer.

\subsubsection{Input Representation \& Positional Encoding On Patient Data}
Given a time-varying covariates sequence of a patient, denoted as \( \bm{x} \), it is transformed as:

\begin{align}
E(\bm{x}) &= \bm{xW}_{\text{emb}} + b_{\text{emb}} \\
P_{\text{pos}, 2i} &= \sin\left(\frac{\text{pos}}{10000^{2i/d_{\text{model}}}}\right) \\
P_{\text{pos}, 2i+1} &= \cos\left(\frac{\text{pos}}{10000^{2i/d_{\text{model}}}}\right) \\
\bm{x}'' &= E(\bm{x}) + P
\end{align}

Given the set of equations, the parameters \( \bm{W}_{\text{emb}} \) and \( b_{\text{emb}} \) are representative of the weight and bias terms for the embedding layer, respectively. Specifically, equation (1) delineates the computation for \( E(\bm{x}) \), which serves as an embedding containing salient information. This embedding is subsequently enhanced with the integration of positional encodings, as defined by the functions \( P_{\text{pos}, 2i} \) and \( P_{\text{pos}, 2i+1} \). The culmination of this process is \( \bm{x''} \), a sophisticated representation encapsulating a patient's comprehensive medical history, amalgamating both categorical and numerical data from their records or visits. Such intricate representations are pivotal in the domain of survival models, where they are employed to prognosticate temporal outcomes related to specific events.

\subsubsection{Transformer Encoder Layers}
After having the enriched representation \( \bm{x}'' \), it will be processed through the Transformer encoder layers \cite{devlin2018bert}. Mathematically, for each layer:
\begin{align}
\text{Attention}(Q, K, V) &= \text{softmax}\left(\frac{QK^T}{\sqrt{d_k}}\right) V
\end{align}
\begin{align}
\bm{x}'''_1 &= \text{LayerNorm}(\bm{x''} + \text{Attention}(Q, K, V))
\end{align}
\begin{align}
    \text{FFN}(\bm{x}) &= \max(0, \bm{xW}_1 + b_1)\bm{W_2} + b_2
\end{align}
\begin{align}
    \bm{x}'''_2 &= \text{LayerNorm}(\bm{x}'''_1 + \text{FFN}(\bm{x}'''_1))
\end{align}


From Eq. (5), the attention scores are computed through the dot product of the Query (Q) and Key (K) matrices, scaled down by a factor of \( \sqrt{d_k} \). These scores dictate the weighting for each component in the Value (V) matrix.  Moreover, the query/key/values in Eq. (7) indeed are from the multi-headed self-attention mechanism of the transformer architecture. In our work, we perform attention-based aggregation across time in our time-varying covariate features described in Section 3.3.2. Moreover, the attention mechanism is particularly adept at capturing and emphasizing the interrelations in a patient's historical events. Moreover, Layer Normalization acts to stabilize the model's activations, whereas the Feed-Forward Network (FFN) introduces necessary non-linearity, as formalized in Eq. (7). The intermediary representation, \( \bm{x}'''_1 \), can be viewed as an enriched perspective of a patient's medical history, vital for subsequent survival analysis tasks. Once processed through the encoder layers, the final output is represented by \( \bm{x}'''_2 \), which encapsulates the combined effects of the initial embedding, attention mechanism, and the FFN transformations.


\subsubsection{Pooling and Feature Extraction}
We employ global average pooling to aggregate the temporal dimension. This is then projected into an \(N\)-dimensional feature vector, aligning it with the SurvTrace input defined as numerical input:

\begin{align*}
y = F(Gl(\bm{x}_{2}''')) = [f_1, f_2, f_3, \ldots, f_{N}] \\
\end{align*}

Where \( Gl \) denotes the global average pooling function, which essentially reduces the spatial dimensions of the input by computing the average over the entire feature map. \( F \) represents the feed-forward transfer function, typically introducing non-linearity and transforming the data for subsequent processing. In addition, originally sourced from individual patient records, the data has now been distilled into \( N \) salient features. This dimensionality reduction ensures computational efficiency and captures the most informative patterns from the patient records.

\subsection{Modified SurvTrace Module}

In SurvTrace \cite{wang2022survtrace}, we have modified its architecture to transition from an initial dual-type input (both numerical and categorical) to exclusively accepting our model's embedding output. This enhancement leverages the Transformer architecture, facilitating not only the management of static covariates as seen in the traditional SurvTrace but also dealing with time-varying covariates data, which contain more meaningful information that can bolster prediction accuracy. In addition, while retaining the beneficial features of the original SurvTrace, our model demonstrates enhanced performance and yields improved results compared to its predecessor.

Moreover, by integrating our method into SurvTrace, we have the capability to enhance its existing loss functions, as detailed as follows:
\[ \mathcal{L} = \mathcal{L}_{\text{IPS}} + \gamma_1 \mathcal{L}_{\text{MP}} + \gamma_2 \mathcal{L}_{\text{LS}}.\]
where \({L}_{LS}\), \({L}_{MP}\) and \({L}_{IPS}\) is the loss for length-of-stay, mortality, and leverage the inverse propensity score, respectively. Two hyper-parameters, \( \gamma_1 \) and \( \gamma_2 \), can initially be set to 1 and subsequently annealed during training.

\begin{table*}
\centering
\caption{Descriptive statistics of three public medical datasets.}
\renewcommand{\arraystretch}{0.9}  
\setlength{\tabcolsep}{6.0pt}      
\begin{tabular}{|l|c|c|c|c|c|c|c|c|c|}
\hline
Dataset & Events (\%) & Censored (\%) & Covariates & \multicolumn{3}{c|}{Event Duration} & \multicolumn{3}{c|}{Censoring Time} \\
\cline{5-10}
& & & (num., cat.) & min & max & mean & min & max & mean \\
\hline
METABRIC & 57.9 & 42.1 & 5, 4 & 0.1 & 355.2 & 99.9 & 0 & 337 & 159.5 \\
\hline
SUPPORT & 68.0 & 32.0 & 8, 6 & 3 & 1944 & 205.4 & 344 & 2029 & 1059.8 \\
\hline
PBC2 & 44.87 & 55.12 & 8, 7 & 41 & 5071 & 2219  & 533  & 5222  & 3368 \\
\hline
\end{tabular}
\end{table*}

\section{Experiments}
\label{sec:experiments}
 In this section, we detail experiments that are aimed at showing the following:
 \begin{enumerate}
 \item Our proposed method ``SurvTimeSurvival'' achieves performance that exceeds state of the art approaches on both covariates and time-varying covariates datasets. 
 \item We provide experimental evidence that the ability to utilise time-varying covariates data within Transformer architectures results in improved predictions of survival curves.
 \item We also show the contribution of the inclusion of synthetic data to our learning process in obtaining the final results for our novel method.
 \end{enumerate}


\subsection{Environment Configuration}
In our experimental framework, we employed the Adam optimization algorithm \cite{kingma2014adam}. Our module was architectured to yield a vector of dimensions ranging from 15 to 30 features, which subsequently served as input to the \textit{SurvTrace} model \cite{wang2022survtrace}. For hyper-parameter tuning, we calibrated learning rates between $1 \times 10^{-4}$ and $1 \times 10^{-3}$, weight decay from $1 \times 10^{-3}$ to $0$, transformer layer depths of 2 to 4, embedding dimensions set at 16, intermediate layers of sizes 32 and 64, and evaluated multi-head attention configurations with heads numbering 1, 2, and 4.

Both the cause-specific and task-specific subnetworks were structured as multi-layer perceptrons (MLPs), encompassing one or two layers. They were dimensionally harmonized with the primary transformers. The ReLU activation functions \cite{schmidt2020nonparametric} were integrated, aligning with the intrinsic design of the \textit{SurvTrace} model.

\subsection{Dataset Setup}


Our study analyzes the event using three different public medical time-varying covariates/ covariates dataset. With covariates data, Study to Understand Prognoses Preferences Outcomes and Risks of Treatment(SUPPORT2) \cite{knaus1995support} and Molecular Taxonomy of Breast Cancer International Consortium (Metabric) \cite{curtis2012genomic} dataset from the Pycox library \cite{kvamme2019continuous} is used. For time-varying covariates data, we employ the Mayo Clinic Primary Biliary Cirrhosis (PBC2) dataset \cite{lindor1996effects} from the Auton library \cite{nagpal2021deep}. Also, our data preparation steps lean heavily not only on our proposed method (Section 3.2) but also on tools from the Auton library \cite{nagpal2021deep}. The statistics of our experimented datasets are available in Table 1.

Regarding synthetic data generation in training/ validation, for SUPPORT2 \cite{knaus1995support} and Metabric \cite{curtis2012genomic} covariates datasets, we employ 'survival\_gan' \cite{qian2023synthcity, norcliffe2023survivalgan} approach. The 'time\_gan' module \cite{yoon2019time, norcliffe2023survivalgan} is employed for training and generating the data for time-varying covariates synthetic data.

Regarding the proportion of data for training/validating and generating, for the covariates datasets, we adopt a strategy wherein 80\% of the original data is used for training and subsequent synthetic data generation, drawing inspiration from the methodology delineated in \cite{gowal2021improving}. When addressing the time-varying covariates, PBC2, we generate 799 synthetic entries pertinent to 188 patients employing the 'time\_gan' model \cite{yoon2019time}. This corresponds to 50\% of the training dataset. The rationale behind this particular proportion for the time-varying covariates dataset is to encapsulate half of the meaningful temporal patterns, thus facilitating an exploration into the model's adaptability and sensitivity to such intricate information dynamics.


 \subsection{Baselines \& Evaluation Metrics}

To assess the efficacy of our proposed method, we conducted benchmark comparisons against state-of-the-art models, including DeepSurv \cite{katzman2018deepsurv}, Deep-Hit \cite{lee2018deephit}, Dynamic Deep-Hit \cite{lee2019dynamic}, and SurvTrace \cite{wang2022survtrace}. For a fair comparison, we obtained the implementation of these methods from their respective authors and applied them to our training and test datasets.
In order to analyze the contribution of synthetic data use in our method, we also perform experiments with and without their inclusion.

We also employed five-fold cross-validation (CV) for a comprehensive evaluation, reporting metrics as the mean values, ensuring precision and uniformity in the comparative analysis. Furthermore, for the \( k \)-th event evaluation, we employed the time-dependent Concordance index \( C_{\text{td}} \) and Brier score:
\begin{align}
    C_{\text{td}}(\tau, k) &= \Pr\left\{ S_k(\tau |\bm{x}_i) > S_k(\tau |\bm{x}_j) \mid e_i \right. \\
    &\quad \left. = k, t_i < t_j, t_i \leq \tau, k > 0 \right\}, \nonumber
\end{align}

\begin{figure*}
  \centering
  \includegraphics[width=0.32\linewidth]{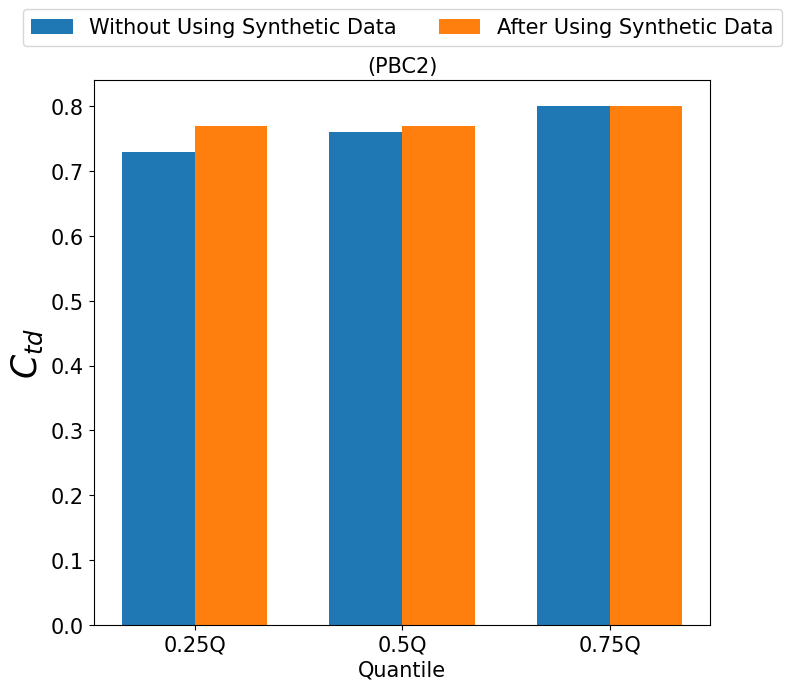}
  \hspace{0.005\linewidth} 
  \includegraphics[width=0.32\linewidth]{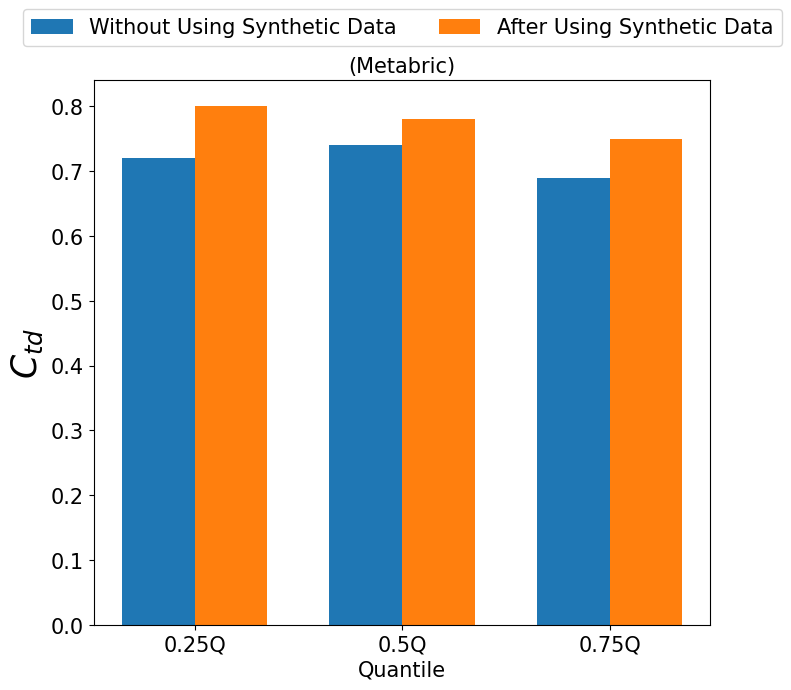}
  \hspace{0.005\linewidth} 
  \includegraphics[width=0.32\linewidth]{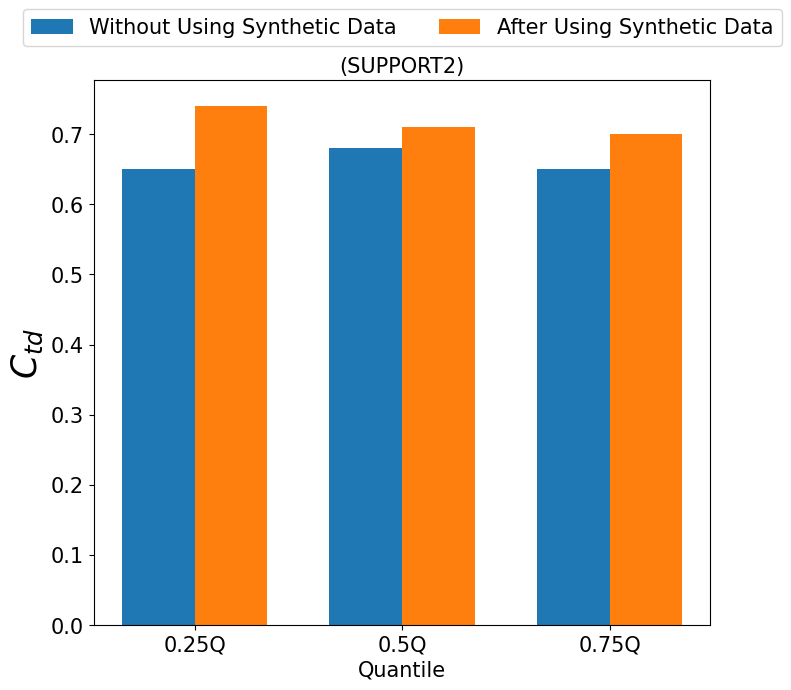}
  \caption{Means of \(C_{td}\) on medical datasets with/without synthetic data on SurvTimeSurvival (CV \(K\)=5)}
  \label{fig:synth_data_ablation}
\end{figure*}

\begin{table*}[h!]
\centering
\caption{Means of \(C_{td}\) for different models on covariates datasets (CV \(K=5\)); The means of standard deviation on five runs shown in the bracket.}
\small 
\setlength{\tabcolsep}{6.5pt} 
\begin{tabular}{|c|c|c|c|c|c|c|}
\hline
 \textbf{Approaches} & \multicolumn{3}{c|}{\textbf{SUPPORT2}} & \multicolumn{3}{c|}{\textbf{METABRIC}} \\
\cline{2-7}
 & \textbf{0.25Q} & \textbf{0.5Q} & \textbf{0.75Q} & \textbf{0.25Q} & \textbf{0.5Q} & \textbf{0.75Q} \\
\hline
DeepSurv & 0.60(0.015) & 0.57(0.011) & 0.59(0.007) & 0.66(0.019) & 0.64(0.025) & 0.64(0.022) \\
\hline
Deep-Hit & 0.62(0.011) & 0.57(0.010) & 0.55(0.012) & 0.70(0.028) & 0.65(0.023) & 0.59(0.023) \\
\hline
Dynamic Deep-Hit & 0.63(0.009) & 0.58(0.010) & 0.55(0.010) & 0.71(0.021) & 0.65(0.015) & 0.61(0.016) \\
\hline
SurvTrace & 0.65(0.010) & 0.62(0.009) & 0.59(0.011) & 0.72(0.023) & 0.71(0.020) & 0.66(0.019) \\
\hline
\cellcolor{gray!30}\textbf{Ours} & \cellcolor{gray!30}\textbf{0.74(0.009)} & \cellcolor{gray!30}\textbf{0.71(0.011)} & \cellcolor{gray!30}\textbf{0.70(0.011)} & \cellcolor{gray!30}\textbf{0.80(0.019)} & \cellcolor{gray!30}\textbf{0.78(0.015)} & \cellcolor{gray!30}\textbf{0.75(0.015)} \\
\hline
\end{tabular}
\end{table*}

\begin{table*}[h!]
\centering
\caption{Means of \(C_{td}\) and Brier score across models on PBC2 time-varying covariates dataset (CV \(K=5\)); The means of standard deviation on five runs shown in the bracket.
}
\small
\setlength{\tabcolsep}{5.5pt}
\begin{tabular}{|c|c|c|c|c|c|c|}
\hline
\multirow{3}{*}{\textbf{Approaches}} & \multicolumn{3}{c|}{\textbf{Concordance Index}} & \multicolumn{3}{c|}{\textbf{Brier Score}} \\
\cline{2-7}
 & \textbf{0.25Q} & \textbf{0.5Q} & \textbf{0.75Q} & \textbf{0.25Q} & \textbf{0.5Q} & \textbf{0.75Q} \\
\hline
DeepSurv & 0.56(0.020) & 0.54(0.026) & 0.55(0.018) & 0.163(0.002) & 0.312(0.004) & 0.322(0.004) \\
\hline
Deep-Hit & 0.59(0.030) & 0.57(0.026) & 0.57(0.024) & 0.127(0.002) & 0.296(0.003) & 0.331(0.003) \\
\hline
Dynamic Deep-Hit & 0.69(0.010) & 0.65(0.009) & 0.64(0.008) & 0.061(0.002) & 0.211(0.003) & 0.223(0.003) \\
\hline
SurvTrace & 0.69(0.015) & 0.64(0.017) & 0.64(0.012) & 0.058(0.002) & 0.210(0.002) & 0.212(0.002) \\
\hline
\rowcolor{gray!30}
Ours - W/o \(G\) & 0.73(0.012) & 0.76(0.010) & 0.80(0.006) & 0.054(0.001) & 0.139(0.002) & 0.189(0.002) \\
\hline
\rowcolor{gray!30}
\textbf{Ours} & \textbf{0.77(0.010)} & \textbf{0.77(0.009)} & \textbf{0.80(0.005)} & \textbf{0.051(0.001)} & \textbf{0.135(0.002)} & \textbf{0.189(0.001)} \\
\hline
\end{tabular}
\label{tab:pbc2_results}
\end{table*}

\begin{align}
    \text{Brier Score} &= \frac{1}{N} \sum_{i=1}^{N} (f_i - o_i)^2.
\end{align}

From (9), \( S_k(t|x_i) \) signifies the  survival function for \( k \)-th event at time \( \tau \). Adjustments were made using IPCW \cite{uno2011c}, and consistent with \cite{nagpal2021deep}, reported \( C_{\text{td}} \) at the 25\%, 50\%, 75\% quantiles. From (10), Brier score with \( N \) predictions, forecasted probability \( f_i \), and actual outcome \( o_i \), the score ranges from 0 (perfect) to 1 (worst), reflecting the squared discrepancy between forecasts and outcomes.


Regarding the experimentation on methods not inherently designed for handling time-varying covariates data \cite{wang2022survtrace, katzman2018deepsurv, lee2018deephit}, we transformed the dataset into a covariates-only dataset. By testing their proposed methods, we highlight the advantages of addressing time-varying covariates over covariates data. 

\subsection{Experiment Results}

The experimental results in terms of \(C_{td}\) values comparing our method ``SurvTimeSurvival'' against SOTA approaches can be seen in Tables 2 \& 3 for the covariates and time-varying covariates datasets respectively.

It can be seen that our methods have performances that exceed every method that we compare against across all datasets.
In the SUPPORT2 dataset, our concordance indices surpassed those of the \textit{SurvTrace} method by margins of 0.09, 0.09, and 0.11 for the 0.25, 0.5, and 0.75 quantiles, respectively. Similarly, for the Metabric dataset, we observed improvements of 0.08, 0.07, and 0.09 across the corresponding quantiles, as described in Table 4. Additionally, compared with SurvTrace, SurvTimeSurvival exhibits advanced capabilities by generating synthetic data and managing time-varying covariates efficiently, highlighting its broad adaptability.

For the PBC2 dataset (Table 3), our method also obtains the highest concordance index values of 0.77, 0.77, and 0.80 across quantiles 0.25, 0.5, and 0.75, respectively. This is an improvement over the next highest performer of Dynamic Deep-Hit of 0.08, 0.12, and 0.16, respectively. Furthermore, considering the Brier Score—a metric to measure the accuracy of probabilistic predictions—our method consistently exhibits commendable results. Specifically, at the 0.25 quantile, our model yielded a score of 0.051, which declined to 0.135 at the 0.5 quantile and remained stable at 0.189 for the 0.75 quantile. Compared with other methods, these scores clearly underscore our approach's efficacy and robustness in predicting survival outcomes, especially compared to the benchmark techniques listed.

\subsubsection{\textbf{Importance of Time-Varying Covariates and Transformers}}

Table \ref{tab:pbc2_results} shows that explicit use of time-varying covariates data can improve prediction accuracy instead of treating each time visit in patient records as discrete. This observation is further supported when contrasting our method with the \textit{SurvTrace} \cite{wang2022survtrace} model when it deals only with covariates data. By capitalizing on the strengths of the Transformer architecture, both our and SurvTrace models register improved performance metrics. Notably, our method outstrips \textit{SurvTrace} by margins of 0.08, 0.13, and 0.16 in the 0.25, 0.5, and 0.75 quantiles, respectively, as evidenced by the Concordance Index since applying time-varying covariates dataset. Moreover, we also instigated a comparison with \textit{Dynamic Deep-Hit} \cite{lee2019dynamic} and recognized that the Transformer architecture has more advantages than the LSTM. 

\subsubsection{\textbf{Contribution of Synthetic Data}}


Table 3 presents the Concordance Index (\(C_{td}\)) comparison of various survival prediction methodologies on the PBC2 dataset. Notwithstanding the absence of synthetic data (w/o \(G\)), our method, \textit{SurvTimeSurvival}, manifests remarkable performance when benchmarked against methods like \cite{katzman2018deepsurv, lee2018deephit, lee2019dynamic, wang2022survtrace}. Remarkably, incorporating synthetic data during the training phase augments the predictive accuracy of \textit{SurvTimeSurvival}. At specified quantiles of 0.25, 0.5, 0.75, \textit{SurvTimeSurvival} enhanced its \(C_{td}\) scores to 0.77, 0.77, and 0.80, respectively, after including the synthetic data. 
The contribution of synthetic data use in our method can be seen in Figure \ref{fig:synth_data_ablation}. We can see an increase in the $C_{td}$ values in all datasets across all quantiles.
Therefore, this experiment highlights the importance of using synthetic data in training, improving accuracy, and potentially addressing situations with limited data due to privacy and technical challenges.

\begin{table}[h!]
\small
    \begin{tabular}{|>{\raggedright}p{4.2cm}|c|c|}
    \hline
    \textbf{Description} & \textbf{SurvTrace} & \textbf{Ours} \\
    \hline
    Apply synthetic data & & $\checkmark$ \\
    \hline
    Dealing w/ "cov" data & $\checkmark$ & $\checkmark$ \\
    \hline
    Dealing w/ "time-cov" data & & $\checkmark$ \\
    \hline
    Dealing w/ multiple events & $\checkmark$ & $\checkmark$\\
    \hline
    Predict survival, risk, hazard & $\checkmark$ & $\checkmark$\\
    \hline
\end{tabular}
\caption{Comparison between SurvTrace \& Our methodology}
\label{tab:comparison}
\end{table}




\section{Conclusion}

In conclusion, the SurvTimeSurvival methodology advances survival analysis by handling both time-varying covariates and covariates data and integrating data synthesis for better accuracy. Building upon the strengths of the Transformer architecture and integrating features from the SurvTrace model, our approach outclasses earlier strategies in precision. It bridges the analytical divide between different data types, promising significant implications not only for healthcare research but also for the engineering and economics domains. In comparison, we find the next highest performing method of SurvTrace only contains a subset of the above capabilities (Table \ref{tab:comparison}). Future directions include integrating knowledge distillation for even greater accuracy.


\bibliography{jmlr-sample}





\end{document}